\documentclass[10pt,logo,copyright]{giga-report}
\usepackage[authoryear,sort&compress,round]{natbib}

\usepackage{iftex}
\ifPDFTeX
    \usepackage[utf8]{inputenc} %
    \usepackage[T1]{fontenc}    %
\fi

\usepackage{parskip}        %
\usepackage{url}            %
\usepackage{booktabs}       %
\usepackage{amsfonts}       %
\usepackage{nicefrac}       %
\usepackage{microtype}      %
\usepackage[dvipsnames]{xcolor} %
\usepackage{graphicx}
\usepackage{animate}        %
\usepackage{subcaption}
\usepackage{tabularx}
\usepackage{makecell}
\usepackage{adjustbox}
\usepackage{setspace}
\usepackage{todonotes}
\usepackage{colortbl}
\usepackage{wrapfig}
\usepackage{svg} 

\definecolor{rankbest}{RGB}{144,238,144}
\definecolor{rankmid}{RGB}{255,255,180}
\definecolor{rankworst}{RGB}{248,150,150}
\definecolor{deltadarkpos}{RGB}{50,180,50}
\definecolor{deltalightpos}{RGB}{180,235,180}
\definecolor{deltayellow}{RGB}{255,255,200}
\definecolor{deltalightneg}{RGB}{250,180,180}
\definecolor{deltadarkneg}{RGB}{220,80,80}

\newcolumntype{M}[1]{>{\centering\arraybackslash}m{#1}}
\usepackage{float}
\usepackage{tikz}
\usetikzlibrary{positioning,shapes,arrows,arrows.meta,fit,tikzmark}
\usepackage{amsmath,amsfonts,bm, bbm,leftindex}
\usepackage{multirow}
\usepackage{comment}
\usepackage{lipsum}
\usepackage[para]{threeparttable}

\def\eqref#1{equation~\ref{#1}}

\def\1{\bm{1}}

\DeclareMathAlphabet{\mathsfit}{\encodingdefault}{\sfdefault}{m}{sl}
\SetMathAlphabet{\mathsfit}{bold}{\encodingdefault}{\sfdefault}{bx}{n}

\makeatletter
\let\save@mathaccent\mathaccent
\newcommand*\if@single[3]{%
  \setbox0\hbox{${\mathaccent"0362{#1}}^H$}%
  \setbox2\hbox{${\mathaccent"0362{\kern0pt#1}}^H$}%
  \ifdim\ht0=\ht2 #3\else #2\fi
  }
\newcommand*\rel@kern[1]{\kern#1\dimexpr\macc@kerna}
\newcommand*\widebar[1]{\@ifnextchar^{{\wide@bar{#1}{0}}}{\wide@bar{#1}{1}}}
\newcommand*\wide@bar[2]{\if@single{#1}{\wide@bar@{#1}{#2}{1}}{\wide@bar@{#1}{#2}{2}}}
\newcommand*\wide@bar@[3]{%
  \begingroup
  \def\mathaccent##1##2{%
    \let\mathaccent\save@mathaccent
    \if#32 \let\macc@nucleus\first@char \fi
    \setbox\z@\hbox{$\macc@style{\macc@nucleus}_{}$}%
    \setbox\tw@\hbox{$\macc@style{\macc@nucleus}{}_{}$}%
    \dimen@\wd\tw@
    \advance\dimen@-\wd\z@
    \divide\dimen@ 3
    \@tempdima\wd\tw@
    \advance\@tempdima-\scriptspace
    \divide\@tempdima 10
    \advance\dimen@-\@tempdima
    \ifdim\dimen@>\z@ \dimen@0pt\fi
    \rel@kern{0.6}\kern-\dimen@
    \if#31
      \overline{\rel@kern{-0.6}\kern\dimen@\macc@nucleus\rel@kern{0.4}\kern\dimen@}%
      \advance\dimen@0.4\dimexpr\macc@kerna
      \let\final@kern#2%
      \ifdim\dimen@<\z@ \let\final@kern1\fi
      \if\final@kern1 \kern-\dimen@\fi
    \else
      \overline{\rel@kern{-0.6}\kern\dimen@#1}%
    \fi
  }%
  \macc@depth\@ne
  \let\math@bgroup\@empty \let\math@egroup\macc@set@skewchar
  \mathsurround\z@ \frozen@everymath{\mathgroup\macc@group\relax}%
  \macc@set@skewchar\relax
  \let\mathaccentV\macc@nested@a
  \if#31
    \macc@nested@a\relax111{#1}%
  \else
    \def\gobble@till@marker##1\endmarker{}%
    \futurelet\first@char\gobble@till@marker#1\endmarker
    \ifcat\noexpand\first@char A\else
      \def\first@char{}%
    \fi
    \macc@nested@a\relax111{\first@char}%
  \fi
  \endgroup
}
\makeatother

\definecolor{darkred}{rgb}{0.7, 0.0, 0.0}

\usepackage{makecell}
\usepackage{tabularx}
\usepackage{graphicx}
\usepackage{booktabs}
\usepackage{multirow}
\usepackage{listings}
\usepackage{wrapfig}
\usepackage{array}
\usepackage{enumitem}
\usepackage[table]{xcolor}
\definecolor{mygray}{gray}{0.9}

\usepackage{amsmath}
\usepackage{pifont}

\usepackage[nameinlink]{cleveref}
\crefname{equation}{Eq.}{Eqs.}
\crefname{figure}{Fig.}{Figs.}
\crefname{section}{Sec.}{Secs.}
\crefname{appendix}{App.}{Apps.}
\crefname{table}{Tab.}{Tabs.}
\crefname{algorithm}{Algo.}{Algos.}
\crefname{thm}{Thm.}{Thms.}
\Crefname{thm}{Thm.}{Thms.}
\crefname{prop}{Prop.}{Props.}

\newcommand{\crefnames}[3]{%
  \@for\next:=#1\do{%
    \expandafter\crefname\expandafter{\next}{#2}{#3}%
  }%
}

\title{GigaWorld-Policy-0.6: Learning Dexterous Manipulation from Human Videos}
\author{
\vspace{-0.1in}
\centerline{GigaAI}
\centerline{Tsinghua University}
\centerline{{Project Page: \href{https://open-gigaai.github.io/giga-world-policy/}{https://open-gigaai.github.io/giga-world-policy/}}}
\footnotesize
\textbf{GigaWorld Team (alphabetical order)}:
\normalfont
Angyuan Ma, Boyuan Wang, Bingyao Yu, Chaojun Ni, Fengnian Zhang, Guan Huang, Guo Li, Guosheng Zhao, Hao Huang, Haoyun Li, Jiwen Lu, Jiangnan Shao, Lingfeng Wang, Mingxuan Ma, Peng Li, Qiuping Deng, Songling Liu, Wentao Tan, Xinze Chen, Xiaojie Jin, Xiaoyu Tian, Xiaofeng Wang, Xinyu Zhou, Yifan Chang, Yanping Fu, Yiquan Gao, Yang Wang, Yun Ye, Zongyan Deng, Zihan Guan, Zhanqian Wu, Zhekai Wang, Zheng Zhu
\vspace{-1em}
}
\begin{document}
\maketitle



\begin{abstract}
Dexterous manipulation is essential for many real-world tasks, yet robots still struggle to achieve human-like agility under occlusion and complex hand–object interactions. World-action models (WAMs) offer a promising approach by jointly learning action generation and visual dynamics. However, robot demonstrations are costly and limited in diversity. Compared with gripper-based manipulation, dexterous manipulation also demands more complex coordination and contact control, resulting in conservative demonstration speeds that imitation-learned policies inherit. We introduce GigaWorld-Policy-0.6, a WAM that addresses these limitations through unified human–robot action representations and temporally compressed policy learning. Our key insight is to express human arm motion and retargeted hand motion in a robot-compatible action space, enabling human videos and robot demonstrations to jointly supervise action generation and visual dynamics. We curate a large-scale egocentric dataset with arm end-effector and dexterous-hand joint trajectories and combine it with real-robot demonstrations for hybrid training, translating diverse human interaction experience into executable robot skills. To address the slow execution of imitation-learned policies, we introduce temporally compressed policy learning, using synchronized frame-strided action and video targets to teach faster, coordinated arm–hand motions while preserving action–visual alignment. Evaluations in simulation and the real world show that GigaWorld-Policy-0.6 ranks first on Bench2Dex and HandsOn,  while achieving
the strongest real-world performance, demonstrating the benefits of human–robot data integration and temporal compression.
\end{abstract}
\abscontent

\section{Introduction}
Dexterous manipulation requires both a broad repertoire of interaction skills and the ability to execute them with agility. Picking up an object, adjusting its pose, and placing it into a confined space involve different arm–hand configurations and changing contact conditions. Humans perform these transitions fluently, coordinating the arm and fingers as the interaction unfolds. Robots, however, often succeed only within a narrow range of demonstrated conditions and execute otherwise simple tasks through prolonged sequences of small motions. Moving toward human-like dexterous manipulation therefore requires policies that can learn from diverse interaction experience and translate that experience into coordinated, efficient execution.

These capabilities are particularly challenging to acquire under partial observability. During manipulation, the hand frequently occludes task-relevant objects and contact regions, while the consequences of an action may become visible only later. A policy must therefore anticipate how the interaction will evolve rather than rely solely on the latest observation. World-action models (WAMs)~\citep{gigaworldpolicy, ma2026openhumanwam, wang2026openwam, cen2026worldbridge, zhou2026drivedreamer, cosmospolicy, ye2026world, wamstep, wang2026reconphys} offer a promising foundation by jointly learning action generation and action-conditioned visual dynamics. However, predictive modeling alone does not determine the breadth or pace of the behaviors a policy learns. These also depend on the interaction experience available for training and the temporal structure of its supervision. For dexterous manipulation, robot demonstrations impose limitations on both.

\begin{figure}[tbp]
    \centering
    \includegraphics[width=\linewidth]{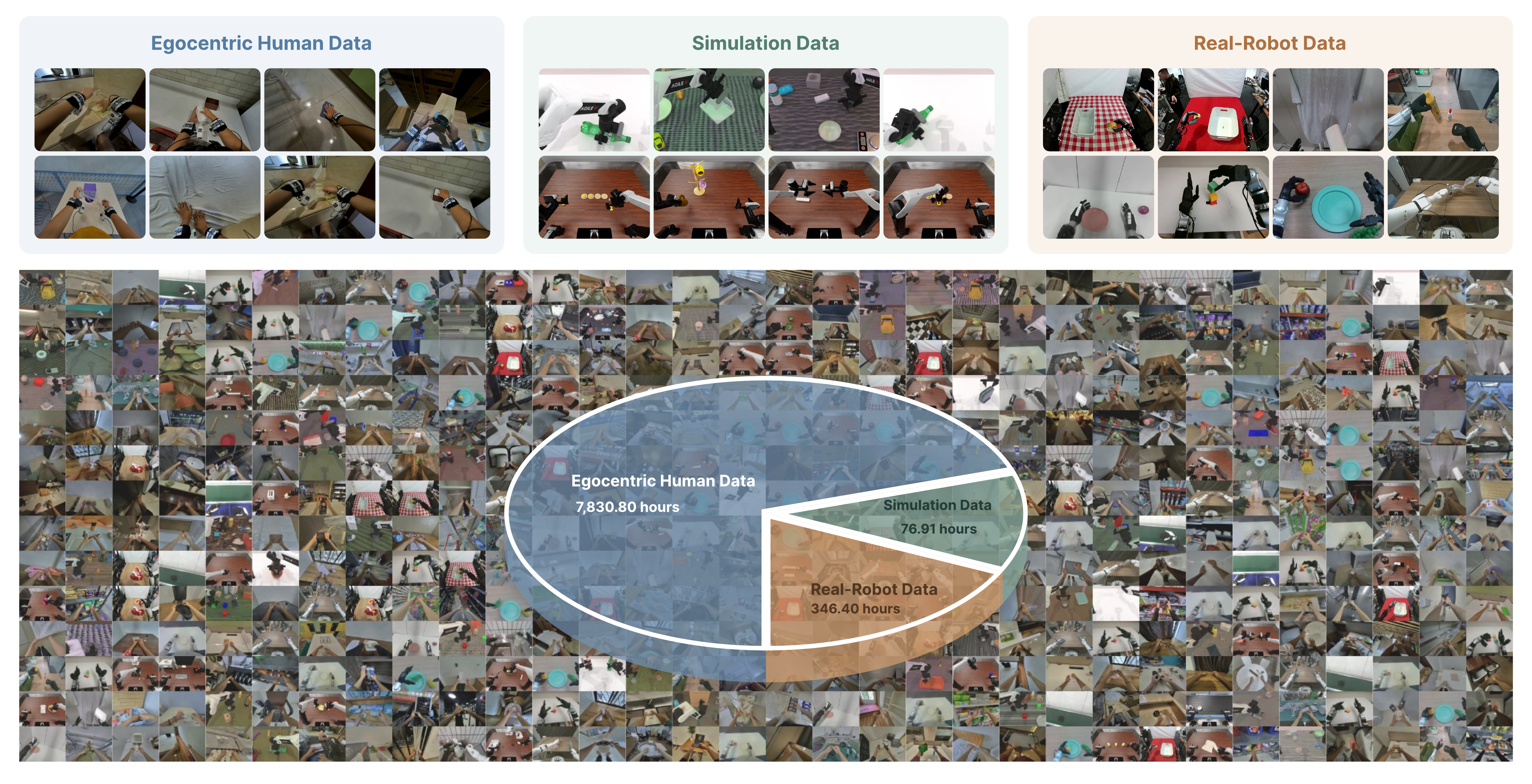}
    \caption{Overview of the pretraining dataset. The dataset combines egocentric human demonstrations, real-robot demonstrations, and simulation trajectories, totaling 7,830.80, 346.40, and 76.91 hours, respectively. It covers diverse manipulation tasks, environments, and robot embodiments.}
    \label{fig:pretraining_data}
\end{figure}

The first limitation is the coverage of manipulation experience. Collecting dexterous robot demonstrations requires substantial effort, making it difficult to capture diverse objects, actions, and interaction states. Egocentric human videos offer a more scalable source of manipulation experience~\citep{yang2025egovla, zheng2026egoscale, hoque2026egodex, wang2026egovid}, but their availability does not directly translate into usable policy supervision. Unlike robot demonstrations, these videos lack executable action labels, and differences in human and robot kinematics prevent observed human motion from being used directly as robot commands. The central challenge is thus to establish a common action representation through which human interaction experience can contribute to robot policy learning.

Our key insight is that human and robot manipulation can be connected through a shared representation of arm motion and robot-compatible hand articulation. We extract arm end-effector trajectories from egocentric human manipulation and retarget human hand motion to dexterous-hand joint trajectories, expressing both in a unified arm–hand action space. This representation turns human videos into action-annotated interaction sequences that can supervise the same action-generation and visual-dynamics objectives as robot demonstrations. We use it to construct a large-scale egocentric manipulation dataset and perform hybrid training with real-robot data, enabling the model to learn shared interaction patterns across human and robot embodiments and translate diverse human experience into executable robot skills.

Broader interaction experience, however, does not by itself ensure agile execution. Imitation-learned policies reproduce the temporal patterns of their demonstrations alongside the demonstrated behaviors. Compared with gripper-based manipulation, dexterous-hand teleoperation requires coordinating more degrees of freedom and managing more complex contacts. Demonstrations are consequently often collected at conservative speeds to maintain control accuracy and hardware safety. A policy can therefore learn the correct manipulation behavior while still executing it through slow, incremental motions. Reducing inference latency addresses the time spent generating actions, but does not directly change the task progression encoded in those actions. Improving physical execution speed requires addressing the temporal structure of the training targets.

To address this limitation, we introduce temporally compressed policy learning. Through synchronized frame-strided sampling, we construct action and video targets that capture greater task progression over a fixed sequence length. Sampling both modalities together preserves the correspondence between arm–hand actions and their visual consequences, allowing the WAM to learn coordinated behavior at a compressed temporal scale. This complements human–robot hybrid training: the unified action representation broadens the manipulation experience available to the policy, while temporal compression teaches the policy to execute learned behaviors more efficiently.

Together, these designs form GigaWorld-Policy-0.6, a WAM for diverse and agile dexterous manipulation. We evaluate our method on simulation benchmarks and real-world manipulation tasks. GigaWorld-Policy-0.6 outperforms the strongest prior method on Bench2Dex by 5.77 percentage points in SR, and leads HandsOn by 1.5 percentage points in SR, while also achieving the strongest aggregate performance across four real-world dexterous tasks.Scaling and ablation studies further examine the complementary benefits of action-annotated egocentric data and temporally compressed policy learning.

Our contributions are threefold:
\begin{itemize}
    \item We present GigaWorld-Policy-0.6, a world--action model for diverse
    and agile dexterous manipulation. It combines scalable interaction
    learning with explicit execution-speed control within a shared policy.

    \item We construct a large-scale action-annotated egocentric corpus with
    arm end-effector trajectories and retargeted dexterous-hand joint
    trajectories. A unified arm--hand action representation enables joint
    training on egocentric human, real-robot, and simulation trajectories,
    transferring diverse human manipulation experience to executable robot
    control.

    \item We introduce temporally compressed policy learning with synchronized frame-strided action and video targets to promote faster, coordinated execution. Evaluations demonstrate that GigaWorld-Policy-0.6 achieves strong task performance in both simulation and real-world settings, while enabling substantially faster physical execution with limited loss in reliability.
\end{itemize}
\section{Related Work}

\subsection{World--Action Models for Robotic Manipulation}
Recent advances in robot learning have benefited from expressive policy representations and large-scale pretraining~\citep{chi2024diffusionpolicy,zhao2023act,brohan2023rt2,kim2024openvla,gigabrain0,team2026gigabrain05,team2026gigaworld,mimicdreamer,embodiedreamer,ni2026swiftvla,ye2026halo, gigaworldpolicy05, wang2026zero2skill}. Building on these advances, World–Action Models (WAMs) integrate predictive world modeling into robot policy learning. Recent approaches adapt pretrained video generative models to model robot action sequences together with future visual observations~\citep{motus,cosmospolicy,ye2026worldactionmodelszeroshot,bi2026motus2}.

Within this paradigm, GigaWorld-Policy~\citep{gigaworldpolicy} and Fast-WAM~\citep{yuan2026fast} show that video prediction can provide useful training supervision while explicit future-video generation can be omitted at inference. OpenWAM~\citep{wang2026openwam} systematically studies WAM architectures, information flow, and pretraining recipes through a modular framework, and develops OpenWAM-$\alpha$ through joint pretraining on egocentric and robot data, with evaluations spanning multiple embodiments, including dexterous hands.

Recent work also explores how human videos can support WAM-based robot learning. Zero-WAM~\citep{zhou2026zerowam} uses human videos as in-context task specifications to guide a causal video-action model in executing unseen manipulation tasks. Skel-WAM~\citep{cai2026skel} aligns human and robot motion through a shared hand-skeleton interface, enabling joint visual and skeletal dynamics learning without robot action labels for human videos. Our work instead explores how WAMs can jointly acquire diverse manipulation skills from human experience and translate them into fluent, coordinated arm–hand execution. Our goal is to move toward natural, human-like dexterous manipulation that combines a broad repertoire of skills with agile execution.

\subsection{Dexterous Manipulation}
Vision--language--action (VLA) models~\citep{pi_0,pi05,ni2026swiftvla,gigabrain0,team2026gigabrain05,gigabrain07,ye2026vla} connect visual and language understanding with dexterous hand control. DexGraspVLA~\citep{zhong2025dexgraspvla} combines a pretrained vision--language planner with a diffusion-based controller for language-guided dexterous grasping. DexVLA~\citep{wen2025dexvla} integrates a diffusion action expert with a vision--language model and uses an embodiment curriculum to support diverse robot embodiments, including dexterous hands. EgoVLA~\citep{yang2025egovla} extends this direction to egocentric human videos, learning wrist and hand actions that are transferred to robots through inverse kinematics, retargeting, and robot-data fine-tuning.

WAM-based approaches further incorporate predictions of how the scene and hand interactions evolve. PointWAM~\citep{park2026pointwam} jointly forecasts scene and hand motion as 3D point trajectories, supports pretraining on human videos, and retargets predicted hand trajectories into robot actions. DexTacWAM~\citep{yuan2026dextacwam} introduces tactile information into a video world model~\citep{gigaworld0,team2026gigaworld,drivedreamer,drivedreamer2,drivedreamer4d,recondreamer,recondreamer++,Recondreamer-rl,wang2024worlddreamer,chang2026mavereconstructingeditable3d, unidrivedreamer, liu2026recondreamersharp} to jointly predict visual and multi-finger contact dynamics for dexterous control. These methods illustrate complementary uses of predictive modeling for dexterous manipulation, from explicit scene--hand geometry to contact evolution.

Our work instead jointly addresses the diversity of dexterous skills and the agility of their execution. GigaWorld-Policy-0.6 uses a unified arm--hand action representation to let human and robot experience jointly supervise action generation and visual dynamics. It combines this supervision with temporally aligned action--video targets and candidate-based speed selection, enabling a shared policy to learn from heterogeneous experience and adapt its execution tempo. This design targets natural, human-like dexterous manipulation through both diverse skill acquisition and fluent, coordinated arm--hand execution.

\subsection{Accelerating Robot Policy Execution}
Several lines of work seek to reduce the time required for a learned policy to complete a task. One strategy changes the temporal structure of demonstrations before or during imitation learning. DemoSpeedup~\citep{demospeedup} applies entropy-guided subsampling to produce faster training trajectories, whereas EgoSpeedUp~\citep{oh2026egospeedup} uses the tempo of human manipulation to retime robot demonstrations. A complementary strategy optimizes speed after imitation learning: SpeedAug~\citep{speedaug} fine-tunes policies trained with speed-augmented demonstrations, while SpeedTuning~\citep{speedtuning} learns speed decisions for a frozen policy. Together, these methods address acceleration through altered training data or an additional optimization stage.

Another direction makes tempo an explicit control variable at deployment. TempoVLA~\citep{tempovla}, for example, trains on retimed demonstrations and uses a vision--language scheduler to choose a speed condition from the current observation before action generation. GigaWorld-Policy-0.6 also supports speed-conditioned execution, but bases online selection on the actions proposed at each available speed. Evaluating these candidates against motion constraints and task context lets the controller account for what the policy is about to do, rather than selecting tempo from the observation alone.

\section{Method}

\subsection{Overview}

\begin{figure}[tbp]
    \centering
    \includegraphics[width=\linewidth]{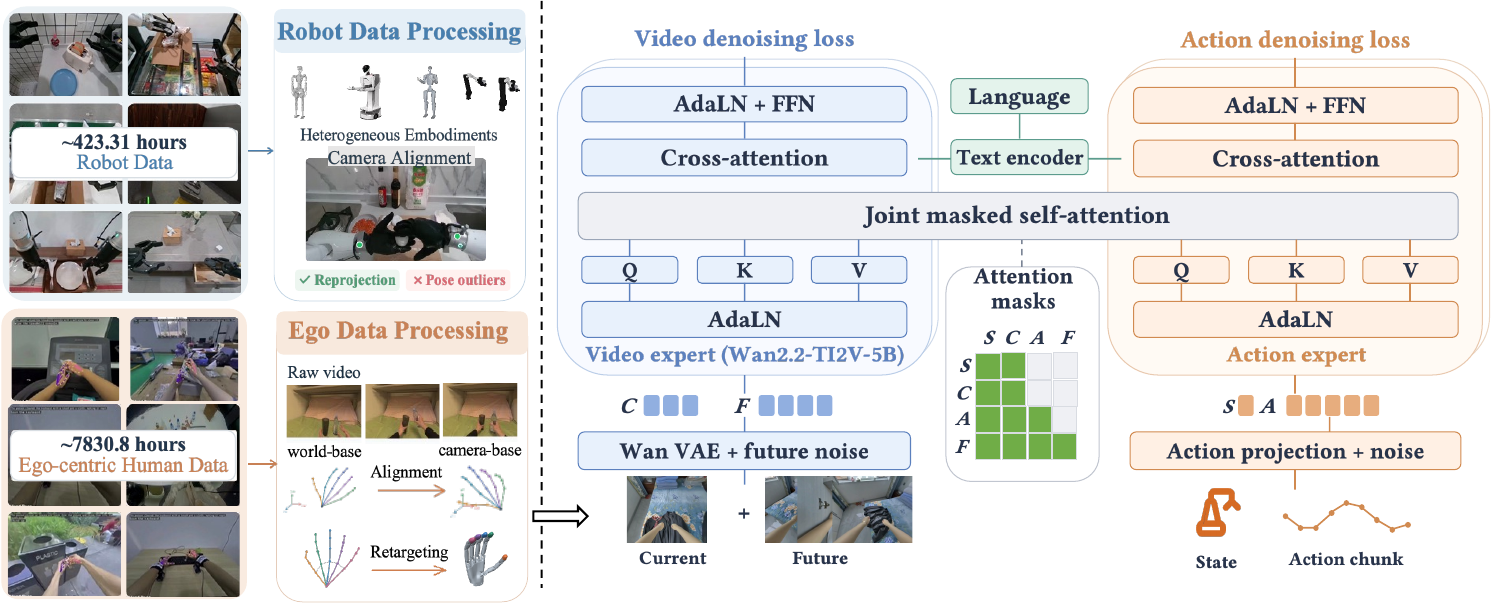}
    \caption{Coordinate-frame alignment and hand-motion retargeting convert egocentric human motion into robot-compatible arm–hand targets, while heterogeneous robot demonstrations are mapped to the shared action space. Conditioned on current observations, embodiment state, and language, a video expert and an action expert jointly learn visual dynamics and action generation through modality-specific denoising objectives. Asymmetric attention allows visual prediction to use action information while preventing action prediction from accessing future visual targets.}
    \label{fig:method}
\end{figure}

We present \textit{GigaWorld-Policy-0.6}, an action-centered world--action model for dexterous manipulation that learns from egocentric human, real-robot, and simulation trajectories, as shown in Figure~\ref{fig:method}. Given current visual observations, a robot state, and a language instruction, the policy predicts a fixed-length executable action chunk. During training, a visual expert additionally learns action-conditioned future dynamics, while a causal attention mask prevents action prediction from accessing future observations.

Our method connects heterogeneous interaction data to agile execution through three components. A unified semantic action representation provides a shared interface for different embodiments and control modalities. An action-centered Mixture-of-Transformers (MoT) jointly learns action generation and visual dynamics through this interface. Temporal scaling then constructs aligned action and video targets at different speeds, supporting both fixed-speed execution and online speed adaptation. We first define the representation and model, then describe data construction and alignment, and finally introduce speed-conditioned training and execution.

\subsection{Action-Centered World--Action Model}

\subsubsection{Unified Action Representation}

\label{sec:unified_action_space}

Our data contain heterogeneous control interfaces, including arm joint commands, Cartesian end-effector targets, parallel grippers, multi-finger hands, mobile-base velocities, and torso motions. To train a single action expert across these modalities, we organize all trajectories into a fixed 94-dimensional semantic action layout. Rather than placing signals according to their order in the original dataset, each action channel is assigned according to its physical meaning.

The first 82 dimensions are divided into two symmetric 41-dimensional blocks for the left and right arms. Each block contains:
(1) seven arm-joint positions or targets,
(2) a three-dimensional end-effector translation,
(3) a six-dimensional continuous rotation representation,
(4) one gripper command, and
(5) 24 dexterous-hand joint targets.
The two arms use the same channel ordering, allowing the shared model to consume both single-arm and bimanual manipulation data. Cartesian rotations are represented by \(\mathrm{rot6d}\) in the model targets; pose annotations stored as quaternions during data processing are converted into this continuous representation when constructing training actions.

Using zero-based indexing, dimensions 82--84 store the mobile-base velocity command \([v_x,v_y,\omega_z]\), and dimension 85 stores the supported torso or lift command. The final eight dimensions are reserved for future embodiment extensions and are not used by the data in this work.

For each trajectory, only the control channels supported by its source embodiment are populated. A dataset recorded in joint space uses the seven arm-joint slots and, when available, the corresponding gripper or dexterous-hand slots. A dataset with reliable Cartesian supervision instead uses the end-effector translation and \(\mathrm{rot6d}\) slots. Parallel-gripper commands are kept distinct from dexterous-finger joints, and the 24 hand channels are assigned following the source hand's finger and thumb joint semantics.

We associate each trajectory with a validity mask $\mathbf{m}\in\{0,1\}^{94}$, where $m_d=1$ indicates a supported action channel. Only valid channels contribute to action supervision. This shared layout preserves embodiment-specific control structures without requiring every source to provide the same control modality.

\subsubsection{Input Representation and Tokenization}

For each training window starting at time $t$, the model is trained on a visual sequence and a future action chunk, conditioned on the current canonical robot state and a language instruction. The action target has length $H$:
\begin{equation}
    \mathbf{a}_{t+1:t+H}
    \in
    \mathbb{R}^{H\times 94},
\end{equation}
and the current state is represented as
\begin{equation}
    \mathbf{s}_t\in\mathbb{R}^{94}.
\end{equation}

The visual input uses three synchronized camera views, including a head view and two auxiliary or wrist-mounted views. The three views are composed into one image frame, with the head view placed in the upper region and the two auxiliary views placed below. We sample visual observations at a temporal stride of $\Delta k$ timesteps, using $K$ future observations to cover the action horizon, so that $H=K\Delta k$:
\begin{equation}
    \mathbf{o}^{\mathrm{comp}}_{t:t+H}
    =
    \left(
        o^{\mathrm{comp}}_t,
        o^{\mathrm{comp}}_{t+\Delta k},
        o^{\mathrm{comp}}_{t+2\Delta k},
        \ldots,
        o^{\mathrm{comp}}_{t+K\Delta k}
    \right).
\end{equation}

The composed frames are encoded by a video VAE into visual latent tokens. The canonical state and action vectors are mapped by dedicated encoders into action-stream tokens. The language instruction is encoded as text conditioning. These tokens form the input to the joint world--action Transformer.

\subsubsection{Action-Centered Mixture-of-Transformers}

GigaWorld-Policy-0.6 uses an action-centered MoT architecture with a visual expert and an action expert.

The visual expert processes current and future visual latent tokens and models how the scene evolves under proposed actions. The action expert processes state and action tokens and performs action-token denoising to predict executable action chunks. The two experts use different hidden dimensions and feed-forward networks, allowing them to specialize for visual dynamics and action generation.

At every Transformer layer, the visual and action experts interact through multimodal self-attention. This design preserves modality-specific expert parameters while enabling information exchange across visual, state, and action tokens. The action expert can use the current visual context, while the visual expert can model scene evolution conditioned on action tokens.

\subsubsection{Joint Flow-Matching Objective}

We train the visual and action branches with flow matching. For each modality, we sample Gaussian noise and a flow-matching timestep, construct a noisy input by interpolating between the clean target and noise, and train the model to predict the corresponding velocity field.

The overall objective is
\begin{equation}
    \mathcal{L}
    =
    \mathcal{L}_{\mathrm{act}}
    +
    \lambda_{\mathrm{vis}}\mathcal{L}_{\mathrm{vis}}.
\end{equation}

Using the validity mask defined above, the action loss is evaluated only on supported dimensions:
\begin{equation}
    \mathcal{L}_{\mathrm{act}}
    =
    \frac{
        \sum_{h=1}^{H}
        \sum_{d=1}^{94}
        m_d
        \left\|
            \hat{v}^{\mathrm{act}}_{h,d}
            -
            v^{\mathrm{act}}_{h,d}
        \right\|_2^2
    }{
        H\sum_{d=1}^{94}m_d
    }.
\end{equation}

The visual loss trains the visual expert to predict the latent evolution of future observations:
\begin{equation}
    \mathcal{L}_{\mathrm{vis}}
    =
    \left\|
        \hat{v}^{\mathrm{vis}}
        -
        v^{\mathrm{vis}}
    \right\|_2^2.
\end{equation}

The two objectives are optimized jointly. The visual branch provides dynamics supervision, while the action branch converts the learned representation into executable robot behavior.

\subsection{From Human Motion to Robot Supervision}

\subsubsection{Data Sources and Composition}

Our pretraining data span three complementary domains: egocentric human trajectories, real-robot trajectories, and simulation trajectories.

The egocentric data are drawn from the Lightwheel EgoSuite~\citep{lightwheel2026egosuite}. The portion used in our pretraining comprises 244{,}741 episodes and 7{,}830.80 hours of manipulation. In addition to RGB observations, these trajectories provide arm end-effector motion and dexterous-hand joint data obtained through hand-pose estimation and retargeting. Therefore, they are not used only for visual pretraining; they provide action signals that can be transformed into robot-policy supervision.

The real-robot data comprise the seven dataset groups listed in Table~\ref{tab:pretraining_data}, collected from different robot setups. These include AgiBot World~\citep{agibot}, ActionNet~\citep{fourier2025actionnet}, DECO-50~\citep{li2026deco}, the Unitree G1 BrainCo PickApple~\citep{unitreebraincoapple} and G1 Dex3 BlockStacking~\citep{unitreedex3stack} datasets, NVIDIA GR00T Teleop G1~\citep{nvidiagr00tg1data}, and T-Rex~\citep{niu2026trex}. They reflect real camera placements, control interfaces, hardware limits, execution latency, and contact dynamics, providing embodiment and physical grounding for the behaviors learned from egocentric data. The simulation data comprise RoboTwin2.0~\citep{robotwin} and RoboDojo~\citep{chen2026robodojo}. RoboTwin2.0 supplies standardized dual-arm tasks, controlled scene variation, and synchronized observation--action pairs.

Together, egocentric data provide broad natural arm--hand interaction priors, real-robot data ground the policy in physical execution, and simulation data provide controllable task coverage. Table~\ref{tab:pretraining_data} summarizes the pretraining datasets.

\begin{table}[t]
\centering
\caption{Statistics for the data used in pretraining. Dataset sources are cited in the Data Sources and Composition section.}
\label{tab:pretraining_data}
\small
\resizebox{0.95\linewidth}{!}{%
\begin{tabular}{lrr|lrr}
\toprule
Dataset & Episodes & Hours & Dataset & Episodes & Hours \\
\midrule
EgoSuite~\citep{lightwheel2026egosuite}                    & 244,741 & 7,830.80 & ActionNet~\citep{fourier2025actionnet} & 32,058 & 156.26 \\
AgiBot World~\citep{agibot}                & 7,203   & 84.09 & DECO50~\citep{li2026deco} & 8,174 & 49.88 \\
G1 BrainCo PickApple~\citep{unitreebraincoapple}        & 200     & 1.42 & G1 Dex3 BlockStacking~\citep{unitreedex3stack} & 301 & 2.60 \\
GR00T Teleop G1~\citep{nvidiagr00tg1data} & 1,095   & 1.72 & TRex~\citep{niu2026trex} & 5,439 & 50.43 \\
RoboTwin2.0~\citep{robotwin} & 27,500  & 56.25 & RoboDojo~\citep{chen2026robodojo} & 3,500 & 20.66 \\
\bottomrule
\end{tabular}
}%
\end{table}


Self-collected post-training data further adapt the shared model to the target real-world tasks. We process each source into the unified representation defined above, aligning coordinates where reliable calibration is available and preserving native control semantics otherwise.

\subsubsection{Ego Data Processing}
\label{sec:ego_processing}

We convert action-annotated egocentric human trajectories into robot-compatible arm--hand supervision. Each trajectory provides egocentric RGB observations, head-camera poses, 21 three-dimensional hand keypoints in MANO ordering, and, when available, task-level language annotations. The objective is to represent human motion in the same action layout used by robot trajectories.

\textbf{Coordinate normalization.} Because the egocentric camera moves with the human head, raw world-frame hand motion contains both hand movement and camera ego-motion. We therefore express all targets in the coordinate frame of the head camera at the current observation. Let \(T^{w}_{c_t}\in SE(3)\) denote the camera-to-world transformation at time \(t\), and let \(P^{w}_{t+k}\in SE(3)\) denote a hand or end-effector pose at a future timestep. The corresponding pose in the current camera frame is
\begin{equation}
P^{c_t}_{t+k} = \left(T^{w}_{c_t}\right)^{-1} P^{w}_{t+k}.
\end{equation}
The same reference frame is used throughout an action chunk, which makes future arm translations and rotations comparable to the observation from which they are predicted. The arm target is represented as an end-effector position together with a unit quaternion, \(\mathbf{a}^{\mathrm{arm}}_t=(\mathbf{p}_t,\mathbf{q}_t)\).

\textbf{Hand retargeting.} Human hand motion is annotated as 21 three-dimensional keypoints per hand, which describe hand geometry but do not directly specify robot-hand joints. We use the retargeting implementation in \texttt{wuji-ego-mint}~\cite{zhu2026mint} to convert these keypoints into joint targets for the Wuji Hand. Retargeting operates in a wrist-centered local hand frame, separating finger articulation from global wrist motion while accounting for the target hand's kinematic structure and joint limits.

For each hand, the resulting 20 joint-angle targets populate the corresponding channels in the shared 24-dimensional hand block. The remaining four channels are marked invalid and excluded from action supervision. Combined with the aligned end-effector trajectories, these targets provide robot-compatible arm--hand supervision for egocentric videos.

We perform quality control on retargeted action trajectories based on geometric consistency, joint constraints, temporal smoothness, and hand-tracking confidence, filtering out kinematically infeasible, discontinuous, and low-confidence samples to ensure training data quality.

\subsubsection{Robot and Simulation Data Alignment}
\label{sec:robot_data_processing}

Our robot and simulation data are collected from heterogeneous embodiments, which differ in camera placement, action interfaces, and state representations. To train a shared policy without mixing incompatible control semantics, we convert each source into a unified action layout while retaining embodiment-specific signals when a common Cartesian representation cannot be reliably established.

\textbf{Coordinate and action-interface normalization.} For datasets that provide the required camera-to-robot or camera-to-base calibration, we transform recorded end-effector poses into the coordinate frame of the current head camera. We apply the same current-camera-frame transformation used for egocentric trajectories. The transformed end-effector target is represented by a three-dimensional translation and a unit quaternion, \((\mathbf{p},\mathbf{q})\), following the same Cartesian representation used for processed egocentric data. This alignment allows human and robot trajectories with reliable pose annotations to share a common arm-action interface.

When accurate calibration or a valid source pose is unavailable, transforming actions into the head-camera frame may introduce artificial motion. In this case, we do not force an EEF representation. Instead, we retain the original joint-space action sequence of that embodiment, which directly corresponds to its control interface and avoids introducing errors caused by unreliable coordinate conversion.

\textbf{Semantic mapping to the unified action space.} After selecting the appropriate action representation, we map source-specific action fields into a shared layout according to their semantic meaning. Cartesian targets occupy the end-effector translation and rotation slots, while native robot commands occupy the corresponding arm-joint slots. Parallel-jaw gripper commands are normalized into a common gripper-openness channel, and dexterous-hand signals are mapped into the finger and thumb slots according to the joint order and coupling rules of the target hand.

This mapping is semantic rather than purely positional: we distinguish end-effector poses from joint commands, gripper opening from finger bending, and independent finger joints from coupled hand-closure signals. The resulting trajectories populate the corresponding semantic slots and validity masks used by the shared model.

\subsection{Temporal Scaling for Agile Execution}
\begin{figure*}[t]
    \centering
    \includegraphics[width=\textwidth]{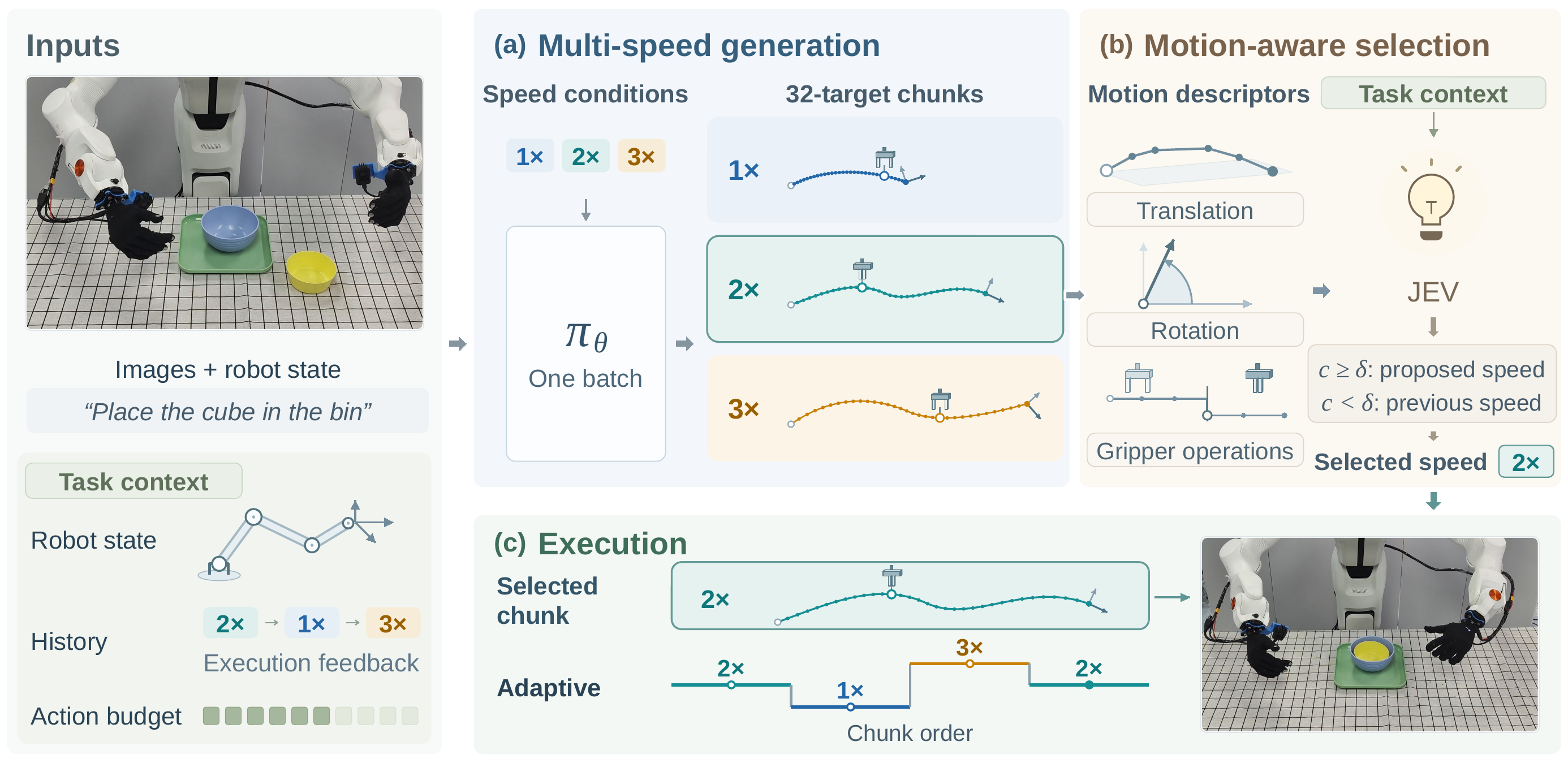}
    \caption{
    (a) One model generates multi-speed action chunks in one batch.
    (b) JEV evaluates motion descriptors using task context
    and proposes a speed condition with confidence $c$.
    If $c<\delta$, the previous speed is retained.
    (c) The robot executes the selected chunk; new observations,
    robot state, and execution history inform the next decision.}
    \label{fig:speed}
\end{figure*}

Scaling the amount of interaction data does not by itself make policy execution agile. Imitation policies reproduce not only the demonstrated motions but also their temporal statistics. If demonstrations are collected with conservative motion and small per-step changes, the learned policy will execute many small increments even when the task could be completed with faster, coordinated motion.

We address this issue through temporal scaling. The key idea is to keep the action chunk length fixed while changing the physical time interval covered by that chunk. This allows one shared policy to represent multiple execution speeds.

\subsubsection{Temporally Aligned Action and Video Targets}

Let $\{(\mathbf{o}_i,\mathbf{a}_i)\}_{i=0}^{L-1}$ denote a synchronized demonstration trajectory after its actions have been converted to the canonical 94-dimensional representation. For a starting index $n$ and a speed condition $s$, we construct an action chunk with $H$ targets:
\begin{equation}
    \mathbf{A}^{(s)}_n
    =
    \left(
        \mathbf{a}_{n+s},
        \mathbf{a}_{n+2s},
        \ldots,
        \mathbf{a}_{n+Hs}
    \right).
\end{equation}

We sample visual targets with the same temporal factor. Using the visual sampling stride $\Delta k$ defined above, let $\mathcal{O}=\{r_0,\ldots,r_K\}$, where $r_j=j\Delta k$ and $K\Delta k=H$. The corresponding sequence is
\begin{equation}
    \mathbf{V}^{(s)}_n
    =\left(\mathbf{o}_{n+s r_0},\ldots,\mathbf{o}_{n+s r_K}\right).
\end{equation}

The action and video targets share the temporal endpoint $n+Hs$. Increasing $s$ therefore covers a longer demonstration interval with the same number of action targets. At a fixed control frequency, these targets represent faster progression through the demonstrated motion. Scaling both modalities together preserves their temporal correspondence for joint action and dynamics training.

\subsubsection{Speed-Conditioned Policy}

To make execution speed explicit, we encode the speed condition with a speed embedding $E(s)$. The embedding is incorporated into the modulation of both the action and visual experts:
\begin{equation}
    m_b(\tau_b,s)
    =
    m_b^{\mathrm{base}}(\tau_b)
    +
    P_b\left(
        \operatorname{SiLU}\left(E(s)\right)
    \right),
    \quad
    b\in\{\mathrm{act},\mathrm{vis}\},
\end{equation}
where $\tau_b$ is the flow-matching timestep, $m_b^{\mathrm{base}}$ is the original timestep modulation, and $P_b$ is a branch-specific projection from the speed embedding to the modulation space.

The same policy parameters are shared across speed conditions. At deployment, the model receives the current observation, canonical state, language instruction, and a requested speed condition, and generates one fixed-length action chunk. No future video needs to be generated during action inference.

For fixed-speed execution, the speed condition remains constant throughout the episode. For adaptive execution, the condition can be updated at action-chunk boundaries according to the current task state.

\subsubsection{Candidate-Based Online Speed Selection}

A speed-conditioned policy provides multiple execution speeds, but the appropriate speed may change during manipulation. Free-space motion can often be executed rapidly, whereas contact formation, alignment, and delicate object handling may require a slower speed. A speed selected in advance may therefore become unsuitable as the task progresses.

We adapt the speed online by comparing candidates generated under different speed conditions. Given the current observation and state, the policy generates a set of action candidates:
\begin{equation}
    \hat{\mathbf{A}}^{(s)}_t
    =\pi_\theta\left(\mathbf{o}_t,\mathbf{s}_t,\ell,s\right),
    \qquad s\in\mathcal{S},
\end{equation}
where $\mathcal{S}$ is the set of available speed conditions. All candidates are generated from the same observation, but each candidate represents a different temporal scale.

The candidates reveal motion information that is unavailable from the speed label alone, including per-step translation, rotation, finger-joint changes, and hand or gripper transitions. The selector compares these proposed motions against the current task state and chooses one candidate for execution.

We consider two selection strategies. The rule-based selector chooses the fastest candidate whose motion characteristics remain within predefined bounds, such as maximum translation, rotation, and hand-state changes. If no high-speed candidate satisfies the constraints, it falls back to a lower speed.

The context-based selector evaluates each candidate using task context and execution history. For candidate $\hat{\mathbf{A}}^{(s)}_t$, it constructs a descriptor that summarizes the proposed motion, hand transitions, distances to interaction points, and expected progress toward the goal. The task context contains the instruction, current object and goal states, remaining task budget, and recently executed speeds.

The selector returns a proposed speed and a confidence score:
\begin{equation}
    \left(\tilde{s}_t,c_t\right)
    =
    \operatorname{Selector}_\rho
    \left(
        \kappa_t,
        \left\{
            \left(s,\phi^{(s)}_t\right)
            \mid s\in\mathcal{S}
        \right\}
    \right),
\end{equation}
where $\kappa_t$ is the task context and $\phi^{(s)}_t$ is the candidate descriptor.

When the confidence reaches a threshold $\delta$, the selected speed is used. Otherwise, the previous speed condition is retained:
\begin{equation}
    s_t =
    \begin{cases}
        \tilde{s}_t, & c_t \geq \delta, \\
        s_{t-1}, & c_t < \delta.
    \end{cases}
\end{equation}

The selected action chunk is then executed by the same low-level controller regardless of its speed condition. After execution, the updated observation, robot state, applied speed, and observed task progress are used to construct the context for the next decision step. This closed-loop design allows the policy to accelerate during safe task-progressing motion while retaining slower execution when contact, alignment, or object stability becomes uncertain.

\section{Experiments}
\label{sec:experiments}

\subsection{Experimental Setup}

\textbf{Evaluation Protocols.}
For simulation, we follow the official Bench2Dex evaluation protocol~\cite{bench2dex}.
Bench2Dex contains 26 task--embodiment settings and four evaluation channels:
\textit{None}, \textit{Equi.}, \textit{Inv.}, and \textit{Full}.
These channels evaluate matched-scene execution, adaptation to task-relevant geometric changes,
robustness to task-irrelevant visual variations, and their combination, respectively.
We conduct 50 rollouts for each task under each channel, resulting in 200 rollouts per task and
5,200 evaluation episodes in total.
We report stable success rate (SR), latched stage completion rate (LSCR), and execution efficiency
(Eff.). LSCR measures the fraction of dependency-consistent task stages reached during execution,
while Eff.\ measures execution efficiency under the benchmark-defined protocol.

\begin{figure*}[t]
    \centering
    \includegraphics[width=\textwidth]{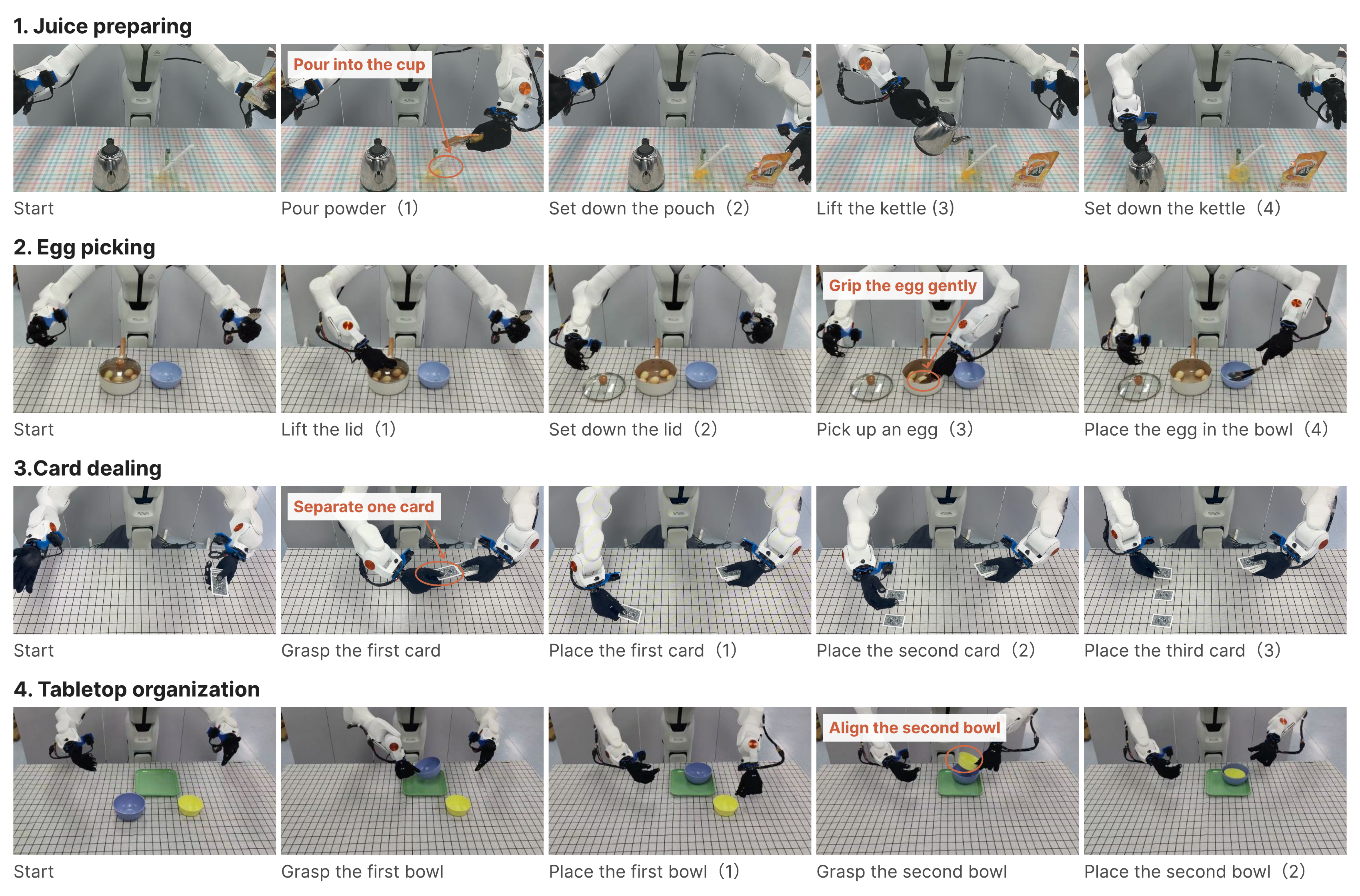}
    \caption{
    Real-world dexterous manipulation tasks and representative execution
    sequences. The text below each task decomposes the task into ordered
    evaluation stages. The value in parentheses denotes the cumulative score
    awarded when execution reaches and completes the corresponding stage.
    }
    \label{fig:real_world_tasks}
\end{figure*}

For real-world evaluation, we consider four dexterous manipulation tasks and conduct 20 trials
per task for each method.
Each task is decomposed into several ordered stages, with predefined scores assigned to completed
stages, as illustrated in Fig.~\ref{fig:real_world_tasks}.
We report the normalized average stage-completion score, referred to as Progress Score (PS), and
the complete-task success rate (SR).
PS gives partial credit when a policy makes valid progress but does not complete the entire task,
whereas a trial contributes to SR only when all required stages are completed.
Both metrics are macro-averaged equally over the four tasks.

\textbf{Baselines.}
On Bench2Dex, we compare GigaWorld-Policy-0.6 with the official baselines spanning three policy
families.
The WAM baseline is Phoenix-MWA~\citep{openphoenix2026phoenix}; the VLA baselines include $\pi_{0.5}$~\citep{pi05}, GR00T N1.5~\citep{nvidia2025gr00tn15}, and GR00T N1.7~\citep{nvidia2026gr00tn17}
with or without tactile observations; and the imitation-learning baselines include ACT~\citep{zhao2023act} and
Diffusion Policy~\citep{chi2024diffusionpolicy}.
For real-world experiments, we compare against $\pi_{0.5}$, Motus, FastWAM, and OpenWAM.
All real-world methods are post-trained using the same task demonstrations and evaluated with
the same trial protocol and initial-condition distribution.

\textbf{Implementation Details.}
GigaWorld-Policy-0.6 is built on the Wan2.2-TI2V-5B diffusion Transformer and uses an action expert adapted to 94-dimensional unified action space.
The model contains 30 Transformer layers with 24 attention heads and an action-expert hidden
dimension of 1,024.
It receives three RGB views, consisting of one head-camera view and two wrist-camera views,
resized to $320 \times 384$.
Unless otherwise specified, we use an action chunk length of $H=32$ and a visual
sampling stride of $\Delta k=8$ timesteps, giving five video frames ($K=4$ future
observations plus the current frame) for visual-dynamics supervision.
Language instructions are represented using precomputed 4,096-dimensional T5 embeddings.
At deployment, we use the action-only inference pathway, which directly denoises action chunks
without decoding future video frames.
Unless otherwise specified, all ablations retain the same architecture and downstream
post-training protocol.

\subsection{Main Results}

\subsubsection{Simulation Benchmarks}

\textbf{Bench2Dex.}
Bench2Dex~\citep{bench2dex} covers 26 bimanual manipulation task--embodiment settings across
12 dexterous embodiments and provides approximately 1.3K teleoperated
demonstration trajectories. We jointly fine-tune a single
GigaWorld-Policy-0.6 model on demonstrations from all training tasks.
We evaluate each setting over 200 episodes, for 5,200 episodes in total.

\begin{table*}[t]
\centering
\caption{
Results on Bench2Dex. Baseline scores are from the official leaderboard~\citep{bench2dexleaderboard} and the Phoenix project report~\citep{openphoenix2026phoenix} (Phoenix-MWA~\citep{openphoenix2026phoenix}).
All methods are evaluated over 26 task--embodiment settings under four evaluation channels.
SR denotes the mean stable success rate across the four channels, LSCR measures
dependency-consistent stage completion, and Eff.\ denotes the benchmark-provided efficiency
score.
The Tactile column indicates whether tactile observations are used.
}
\label{tab:bench2dex_results}
\small
\setlength{\tabcolsep}{5.5pt}
\resizebox{0.95\linewidth}{!}{%
\begin{tabular}{lccccccccc}
\toprule
Method & Type & Tactile & None & Equi. & Inv. & Full & SR & LSCR & Eff. \\
\midrule
ACT~\citep{zhao2023act}
& IL
& --
& 29.30 & 15.50 & 14.80 & 12.80
& 18.10 & 34.90 & 0.967 \\

Diffusion Policy~\citep{chi2024diffusionpolicy}
& IL
& --
& 8.90 & 3.00 & 3.20 & 3.40
& 4.60 & 18.30 & 0.865 \\

$\pi_{0.5}$~\citep{pi05}
& VLA
& --
& 35.20 & 21.20 & 26.80 & 26.20
& 27.40 & 44.70 & 0.960 \\

GR00T N1.5~\citep{nvidia2025gr00tn15}
& VLA
& --
& 38.00 & 16.30 & 24.30 & 20.80
& 24.90 & 45.20 & 0.930 \\

GR00T N1.5~\citep{nvidia2025gr00tn15}
& VLA
& \checkmark
& 38.50 & 17.20 & 23.30 & 22.80
& 25.50 & 46.50 & 0.946 \\

GR00T N1.7~\citep{nvidia2026gr00tn17}
& VLA
& --
& 39.38 & 13.15 & 28.46 & 27.92
& 27.23 & 45.50 & 0.940 \\

GR00T N1.7~\citep{nvidia2026gr00tn17}
& VLA
& \checkmark
& 44.23 & 14.54 & 30.69 & 29.15
& 29.65 & 48.37 & 0.950 \\

Phoenix-MWA~\citep{openphoenix2026phoenix}
& WAM
& --
& 54.20 & 38.50 & 44.10 & 45.60
& 45.60 & 65.60 & -- \\
\midrule
\textbf{GigaWorld-Policy-0.6}
& WAM
& --
& \textbf{55.38}
& \textbf{48.23}
& \textbf{51.38}
& \textbf{50.46}
& \textbf{51.37}
& \textbf{70.91}
& 0.962 \\
\bottomrule
\end{tabular}
}
\end{table*}

As shown in Table~\ref{tab:bench2dex_results}, GigaWorld-Policy-0.6 achieves
the strongest aggregate performance, with an SR of 51.37\% and an LSCR of
70.91\%. Relative to Phoenix-MWA~\citep{openphoenix2026phoenix}, the strongest prior method, this corresponds
to gains of 5.77 and 5.31 percentage points, respectively. The improvement
in both metrics is notable: SR measures stable task completion, while LSCR
captures dependency-valid task stages reached during an episode. Thus, the
gain reflects both more completed tasks and stronger progress through
multi-stage interactions across heterogeneous embodiments. The model achieves
these results without tactile input, with an Eff. score of 0.962.

\textbf{HandsOn.}
We evaluate GigaWorld-Policy-0.6 on HandsOn, a large-scale simulation
benchmark for generalizable dexterous manipulation. HandsOn covers 66 tasks
organized into Core-50, which supports demonstration-based policy learning
and evaluation, and Transfer-16, which assesses generalization to held-out
tasks. It contains 18,000 curated demonstrations, split into 15,178 training and 2,822 validation
trajectories. We report results on Core-50, which comprises 17 primitive,
18 functional, and 15 contact-rich tasks. Each task is evaluated over
100 episodes, for 5,000 episodes in total.

\begin{table*}[t]
\centering
\caption{
Results on the HandsOn Core-50 benchmark. Baseline scores are from Table~5 of HandsOn~\citep{luo2026handson}. SR is task-macro success rate;
Tasks $\geq 50\%$ counts tasks whose success rate is at least 50. P, F, and
CR are macro success rates on the Primitive (17 tasks), Functional (18 tasks),
and Contact-Rich (15 tasks) subsets, respectively. Steps/plan is the number
of 30-Hz control steps generated per policy query. The single-task DP row is
reported for context. Its matched-protocol result covers 11 tasks.
}
\label{tab:handson_core50}
\small
\setlength{\tabcolsep}{7pt}
\resizebox{\textwidth}{!}{%
\begin{tabular}{lcccccc}
\toprule
Model & Steps/plan & SR (\%) $\uparrow$ & Tasks $\geq 50\%$ & P (\%) $\uparrow$ & F (\%) $\uparrow$ & CR (\%) $\uparrow$ \\
\midrule
GPT-6 (move\_eef) & 1--300 & 5.2 & 1 & 7.6 & 1.1 & 7.3 \\
DP~\citep{chi2024diffusionpolicy} (single-task)$^{\dagger}$ & -- & 9.7 & 3 & \multicolumn{3}{c}{Reference: 9.9\% on 11 matched tasks} \\
Spatial Forcing~\citep{li2026spatialforcing} & 16 & 12.4 & 3 & 17.9 & 8.5 & 10.9 \\
Motus~\citep{motus} & 16 & 18.1 & 5 & 30.1 & 11.5 & 12.3 \\
HY-Embodied~\citep{zhang2026hyembodied} & 30 & 18.9 & 8 & 27.1 & 13.3 & 16.4 \\
LingBot-VLA~\citep{wu2026lingbotvla} & 30 & 19.7 & 8 & 26.9 & 14.3 & 17.9 \\
DM0.5~\citep{yu2026dm0,dexmal2026dm05} & 30 & 21.4 & 9 & 37.7 & 10.0 & 16.7 \\
$\pi_{0.5}$~\citep{pi05} & 30 & 24.9 & 10 & 40.4 & 14.6 & 19.7 \\
FastWAM~\citep{yuan2026fast} & 16 & 26.4 & 9  & 42.7 & 20.2 & 15.3 \\
OpenWAM~\citep{wang2026openwam} & 16 & 29.1 & 11 & 40.8 & \textbf{23.8} & 22.1 \\
\midrule
\textbf{GigaWorld-Policy-0.6 (Ours)} & 16 & \textbf{30.6} & \textbf{13} & \textbf{44.5} & 22.7 & \textbf{24.3} \\
\bottomrule
\end{tabular}
}
\begin{minipage}{\textwidth}
\footnotesize $^{\dagger}$The single-task DP aggregate (9.7\%) covers all 50 tasks across different evaluation protocols. On the 11 matched tasks, DP achieves 9.9\%.
\end{minipage}
\end{table*}

As shown in Table~\ref{tab:handson_core50}, GigaWorld-Policy-0.6 achieves
the highest overall success rate (30.6\%) among the compared methods, with
13 tasks reaching at least 50\% success. Compared with the strongest prior
overall result, this is a 1.5 percentage-point gain and two additional tasks
above the 50\% threshold. The gains are particularly notable on the
Primitive and Contact-Rich subsets, where our method exceeds the strongest
prior results by 1.8 and 2.2 percentage points, respectively, using 16
control steps per query. The improvement on Contact-Rich tasks suggests that
the overall result also reflects stronger performance in interaction-heavy
settings.

\subsubsection{Real-World Dexterous Manipulation}
\label{sec:real_world_experiments}
We further evaluate GigaWorld-Policy-0.6 on four real-world dexterous
manipulation tasks: \textit{Tabletop Organization}, \textit{Card Dealing},
\textit{Fruit Juice Preparation}, and \textit{Egg Picking}. These tasks cover
multi-object rearrangement, manipulation of thin objects,
long-horizon sequential execution, and contact-sensitive grasping. We compare
against $\pi_{0.5}$, Motus, FastWAM, and OpenWAM. Each method is evaluated over
20 trials per task using the protocol described in Sec.~\ref{sec:experiments}.
All methods are executed at their nominal speed in this comparison.

\begin{figure*}[t]
    \centering
    \includegraphics[width=\textwidth]{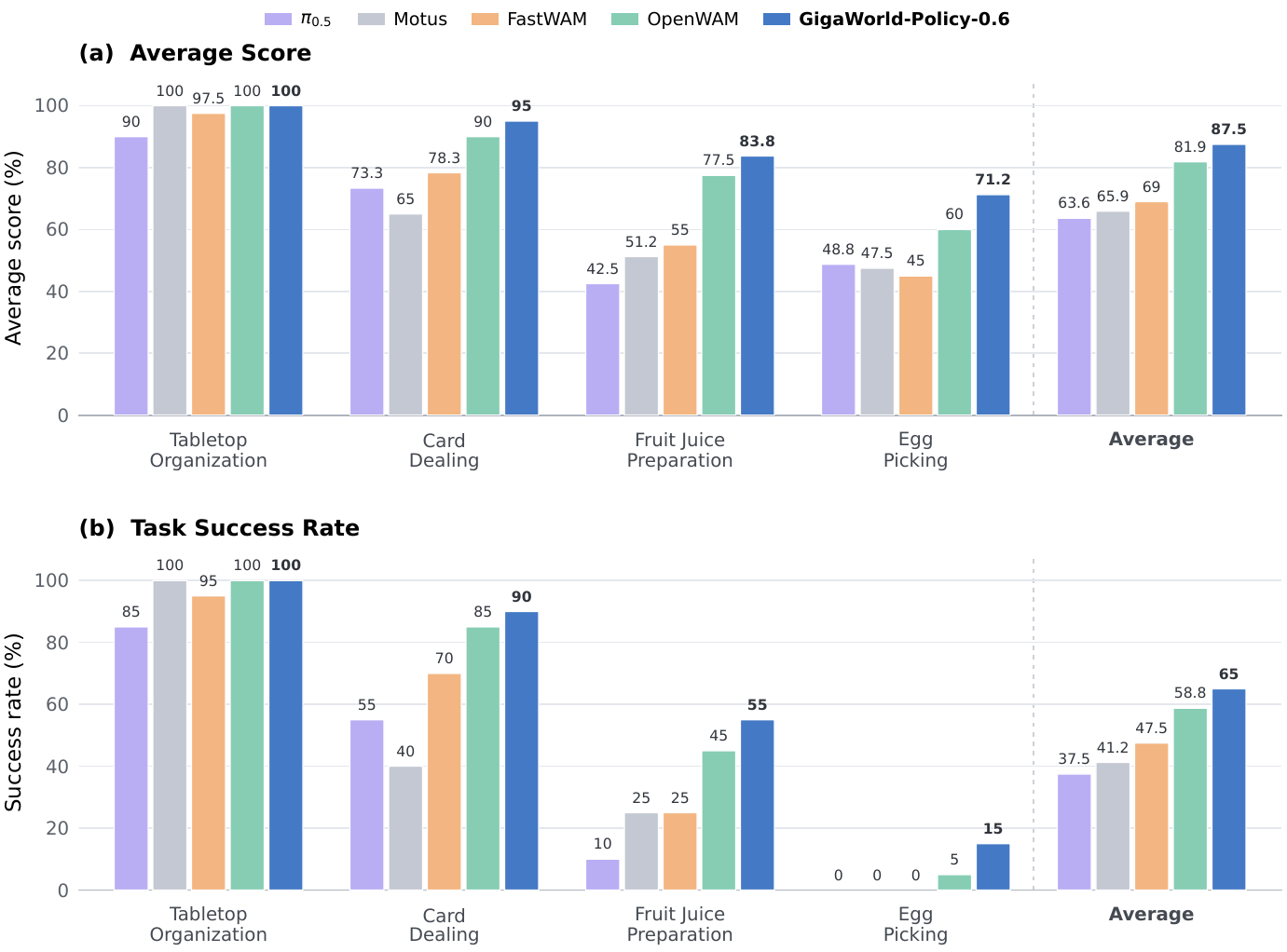}
    \caption{
    Real-world evaluation across four dexterous manipulation tasks, with
    20 trials per task. (a) Progress Score (PS), computed from normalized
    stage completion. (b) Success rate (SR), based on complete task execution.
    }
    \label{fig:real_world_results}
\end{figure*}

As shown in Fig.~\ref{fig:real_world_results}, GigaWorld-Policy-0.6 achieves
the best aggregate performance, with an average PS of 87.5\% and an average SR
of 65.0\%. Compared with the strongest baseline, OpenWAM, our method improves
PS by 5.6 percentage points and SR by 6.3 percentage points. The two methods
perform comparably on tabletop organization, where performance is nearly
saturated, while our model consistently improves both metrics on card dealing,
fruit-juice preparation, and egg picking. The gains are most pronounced on the
last two tasks, which require long-horizon arm--hand coordination and
contact-sensitive grasping, indicating that the improvement extends beyond
coarse object rearrangement to more demanding dexterous interactions.

\subsection{What Does Action-Annotated Ego Experience Contribute?}
\label{sec:ego_analysis}

The main results establish the overall manipulation performance of our model.
We next isolate how egocentric visual experience, arm-motion supervision, and
retargeted hand supervision contribute to this performance, and examine how
their benefits change with pretraining and downstream robot-data scale.

\subsubsection{Contributions of Ego Video and Action Supervision}
\label{sec:pretraining_ablation}

We first separate the contributions of the different supervision signals in
our egocentric corpus. Starting from a model trained without ego data, we add
ego videos alone, then arm EEF targets, and finally retargeted dexterous-hand
joint targets. This progression distinguishes visual-dynamics pretraining from
human arm-motion learning and fine-grained hand-action learning. We additionally
remove real-robot pretraining data to measure the embodiment grounding supplied
by heterogeneous robot trajectories. Simulation pretraining data, model
architecture, optimization budget, and downstream post-training data are held
fixed across all variants.

\begin{table*}[t]
\centering
\caption{
Real-world ablation of pretraining data and egocentric supervision. Ego-V,
Ego-EEF, and Ego-Hand denote ego video, arm EEF targets, and retargeted
dexterous-hand joint targets, respectively. Robot denotes heterogeneous
real-robot pretraining data. PS and SR are macro-averaged over the four tasks.
}
\label{tab:pretraining_ablation}
\small
\setlength{\tabcolsep}{7pt}
\begin{tabular}{lcccccc}
\toprule
Variant
& Ego-V
& Ego-EEF
& Ego-Hand
& Robot
& PS (\%) $\uparrow$
& SR (\%) $\uparrow$ \\
\midrule
w/o Ego Pretraining
& -- & -- & -- & \checkmark
& 69.4 & 50.0 \\

Ego Video Only
& \checkmark & -- & -- & \checkmark
& 70.5 & 52.5 \\

Ego Video + EEF
& \checkmark & \checkmark & -- & \checkmark
& 76.8 & 56.3 \\

w/o Real-Robot Pretraining
& \checkmark & \checkmark & \checkmark & --
& 84.9 & 63.8 \\
\midrule
Full Model
& \checkmark & \checkmark & \checkmark & \checkmark
& \textbf{87.5} & \textbf{65.0} \\
\bottomrule
\end{tabular}
\end{table*}

Table~\ref{tab:pretraining_ablation} suggests that the key benefit of
egocentric pretraining does not come from additional visual observations
alone, but from turning human interaction experience into actionable
supervision. Ego video produces only a limited improvement, indicating that
visual-dynamics learning by itself is insufficient to recover the
fine-grained control structure required for dexterous manipulation. EEF
annotations provide a stronger bridge between human motion and robot control,
allowing the model to learn transferable reaching and arm--hand coordination
patterns. The largest gain comes from adding retargeted hand-joint
supervision, which improves PS and SR by 10.7 and 8.8 percentage points over
the video-and-EEF variant. This result indicates that finger-level motion,
rather than visual diversity alone, is a principal bottleneck in transferring
egocentric experience to dexterous robot policies.

Real-robot pretraining provides a smaller but consistent additional gain,
suggesting a complementary role: ego data supplies broad interaction and
dexterity priors, whereas robot data grounds these priors in embodiment-specific
kinematics, observations, and control dynamics. Together, the results show
that scalable human experience becomes substantially more useful when its
arm and hand motions are explicitly aligned with the robot action space.

\subsubsection{Scaling Egocentric Pretraining}
\label{sec:ego_scaling}

We study how downstream manipulation performance changes as action-annotated
egocentric pretraining is scaled. We construct nested subsets containing
$0\%$, $25\%$, $50\%$, $75\%$, and $100\%$ of the complete ego corpus,
corresponding to approximately 0, 1,958, 3,915, 5,873, and 7,831 hours of human
interaction. Both ego videos and their associated EEF and retargeted
hand-action annotations are retained in every nonzero subset.

Real-robot and simulation pretraining data are kept fixed across all variants.
We use the same number of pretraining optimization steps, data-source sampling
ratios, model architecture, and downstream post-training demonstrations.
Smaller ego subsets are repeatedly sampled when necessary to maintain a
matched optimization budget. Each variant is evaluated on the same four tasks
under the real-world protocol described above.

\begin{figure*}[t]
    \centering
    \includegraphics[width=\textwidth]{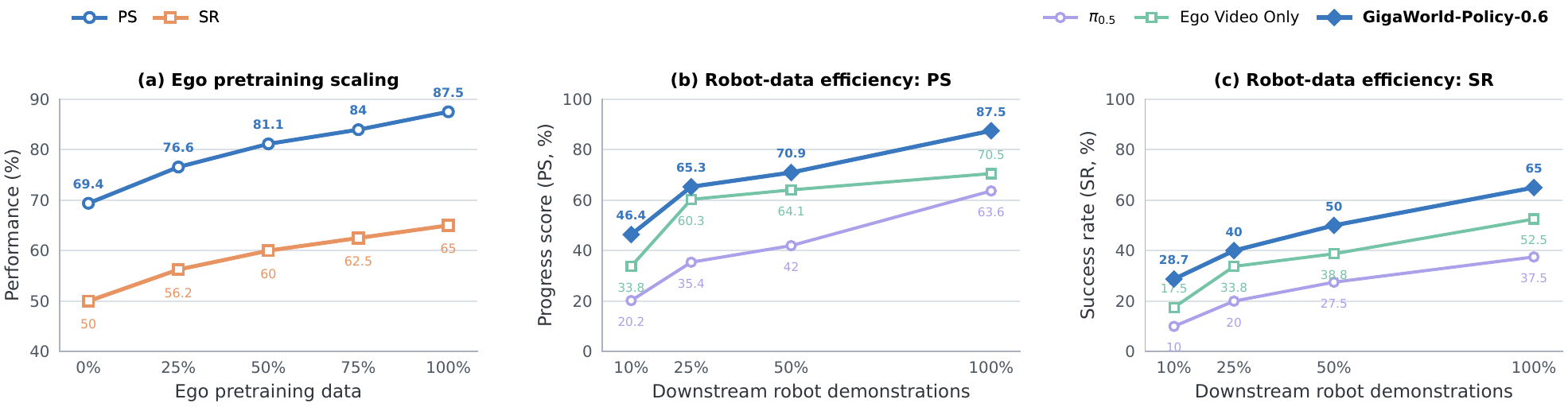}
    \caption{
    Scaling and data efficiency in real-world dexterous manipulation. All
    results report macro-averaged Progress Score (PS) and Success Rate (SR)
    over four tasks. (a) Performance as action-annotated egocentric
    pretraining data increase from $0\%$ to $100\%$, with real-robot and
    simulation pretraining held fixed. (b--c) PS and SR under different
    fractions of downstream robot demonstrations.
    }
    \label{fig:ego_scaling_efficiency}
\end{figure*}

Figure~\ref{fig:ego_scaling_efficiency}(a) shows a clear positive scaling
trend: both PS and SR improve at every increase in egocentric pretraining data.
The first $25\%$ already provides substantial gains, while performance
continues to rise through the full 7,831-hour corpus rather than saturating at
an intermediate scale. This trend is especially meaningful because robot data,
optimization budget, and downstream demonstrations are unchanged. The gains
therefore reflect broader transferable arm--hand interaction experience, not
additional embodiment-specific supervision.

\subsubsection{Robot-Data Efficiency}
\label{sec:data_efficiency}

Collecting embodiment-specific dexterous robot demonstrations is expensive.
We therefore investigate whether action-annotated ego pretraining reduces the
amount of downstream robot data required for adaptation. For each task, we
construct nested subsets containing $10\%$, $25\%$, $50\%$, and $100\%$ of
the available demonstrations. Every method is independently post-trained on
the same subset at each scale using a matched optimization budget, and each
checkpoint is evaluated over 20 trials per task with matched initial
configurations.

We compare $\pi_{0.5}$ with two controlled variants of our model: ego
video-only pretraining and joint ego video--action pretraining. This comparison
separates the benefit of visual experience from the additional contribution of
human arm--hand action annotations.

Figures~\ref{fig:ego_scaling_efficiency}(b--c) show that the advantage of
action-annotated ego pretraining is largest when robot demonstrations are
scarce. With only $25\%$ of the downstream data, GigaWorld-Policy-0.6 reaches
65.3\% PS and 40.0\% SR, already surpassing $\pi_{0.5}$ trained with the
complete robot dataset. It therefore matches the strongest VLA setting using
four times fewer embodiment-specific demonstrations. Ego video-only
pretraining also consistently improves over $\pi_{0.5}$, but remains below
the full model at every data scale. This gap shows that the data-efficiency
gain cannot be attributed to additional visual pretraining alone: aligning
human EEF and dexterous-hand motions with the robot action space provides
actionable priors that are particularly valuable in the low-data regime.

\subsection{How Can the Policy Execute Faster without Sacrificing Reliability?}
\label{sec:speed_analysis}

The preceding experiments evaluate whether the policy can complete dexterous
tasks. We now study whether it can execute the learned behaviors more quickly
without substantially degrading task progress or completion. We first isolate
the effect of mixed-speed training and then compare fixed-speed execution with
online adaptive speed selection.

\subsubsection{Learning Multiple Execution Speeds}
\label{sec:mixed_speed_ablation}

We construct multiple temporal views of each training trajectory through
temporal subsampling and align the corresponding action and future-video
targets to a fixed prediction horizon. A shared speed-conditioned policy is
then trained on the resulting $1\times$, $2\times$, and $3\times$ targets.
To determine whether faster behavior is learned rather than obtained by merely
discarding actions at deployment, we compare mixed-speed training with a
nominal-speed-only policy evaluated using naive temporal action skipping. Both
policies use the same architecture, source trajectories, and optimization
budget.

\begin{table}[t]
\centering
\caption{
Execution-speed control on Tabletop Organization. Naive $2\times$ and
$3\times$ apply temporal action skipping to a policy trained only at nominal
speed, whereas fixed and adaptive mixed-speed settings use the same
speed-conditioned policy. PS and SR are evaluated over 20 trials, and
$T^{\mathrm{real}}$ is the mean completion time over successful trials.
Speedup is measured relative to the corresponding $1\times$ policy.
}
\label{tab:speed_control}
\small
\setlength{\tabcolsep}{5pt}
\begin{tabular}{llcccc}
\toprule
Training & Deployment
& PS (\%) $\uparrow$
& SR (\%) $\uparrow$
& $T^{\mathrm{real}}$ (s) $\downarrow$
& Speedup $\uparrow$ \\
\midrule
Nominal-only & $1\times$
& 100.0 & 100.0 & 82.35 & $1.00\times$ \\
Nominal-only & Naive $2\times$
& 87.5 & 80.0 & 52.59 & $1.57\times$ \\
Nominal-only & Naive $3\times$
& 67.5 & 50.0 & 39.48 & $2.09\times$ \\
\midrule
Mixed-speed & $1\times$
& 100.0 & 100.0 & 81.58 & $1.00\times$ \\
Mixed-speed & $2\times$
& 95.0 & 90.0 & 54.19 & $1.51\times$ \\
Mixed-speed & $3\times$
& 92.5 & 90.0 & 40.38 & $2.02\times$ \\
Mixed-speed & JEV-based
& 97.5 & 95.0 & 47.57 & $1.71\times$ \\
\bottomrule
\end{tabular}
\end{table}

Table~\ref{tab:speed_control} shows that mixed-speed training preserves nominal
behavior while learning substantially more reliable fast motions. At
$1\times$, the two policies both achieve perfect PS and SR with nearly
identical completion times, indicating that mixed-speed supervision does not
degrade normal execution. At $2\times$, mixed-speed training improves PS and
SR over naive skipping by 7.5 and 10.0 points while adding only 1.60 seconds.
The gap widens at $3\times$: mixed-speed execution improves PS by 25.0 points
and SR by 40.0 points with a completion-time difference below one second.
Thus, the benefit does not come from executing more slowly; temporal
subsampling teaches the policy coherent fast trajectories that preserve
critical contact and transition states that naive action removal discards.

\subsubsection{Fixed and Adaptive Speed Control}
\label{sec:adaptive_speed}

A speed-conditioned policy supports multiple execution modes, but a fixed
speed may be inappropriate throughout an entire task. Free-space reaching can
often be accelerated safely, whereas grasp formation, object contact, and
release may require slower and more precise motion. We therefore compare fixed
execution at $1\times$, $2\times$, and $3\times$ with JEV-based online speed
selection using the same mixed-speed-trained policy.

The fixed-speed results in Table~\ref{tab:speed_control} expose the expected
reliability--efficiency trade-off: $3\times$ execution reaches a $2.02\times$
speedup but reduces PS and SR relative to $1\times$. JEV-based selection moves
beyond this fixed trade-off. It reaches a $1.71\times$ speedup while retaining
97.5\% PS and 95.0\% SR. In particular, it is both faster and more reliable
than fixed $2\times$, and it recovers 5.0 PS and 5.0 SR points over fixed
$3\times$ while remaining substantially faster than nominal execution. This
indicates that candidate evaluation uses task context to accelerate
unconstrained motion and preserve precision around contact-sensitive stages.

\section{Conclusion}
We present GigaWorld-Policy-0.6, a world--action policy for general-purpose
dexterous manipulation that learns from large-scale action-annotated
egocentric experience together with heterogeneous robot and simulation data.
A unified action space aligns human arm--hand trajectories and diverse robot
embodiments, allowing egocentric video to provide not only visual priors but
also direct supervision for coordinated arm and dexterous-hand control.
Furthermore, mixed-speed training equips a shared policy with controllable
execution speeds, while adaptive selection improves task efficiency without
requiring a separate policy for each speed. Experiments on Bench2Dex, HandsOn,
and four real-world dexterous tasks demonstrate strong performance across
heterogeneous embodiments, contact-rich interactions, and long-horizon task
stages. Our ablations further show that action annotations are central to
transferring egocentric experience to dexterous control, that performance
continues to improve with increasing ego-data scale, and that robot data
provides complementary embodiment grounding. Together, these results suggest
that scaling action-aligned human experience and explicitly learning execution
agility provide a practical path toward dexterous robot policies that are both
general and efficient.

\clearpage

\setcitestyle{numbers}
\bibliographystyle{plainnat}
\bibliography{./main}

\end{document}